# Knowledge Discovery System For Fiber Reinforced Polymer Matrix Composite Laminate

Doreswamy

**Abstract**—In this paper Knowledge Discovery System (KDS) is proposed and implemented for the extraction of knowledge-mean stiffness of a polymer composite material in which when fibers are placed at different orientations. Cosine amplitude method is implemented for retrieving compatible polymer matrix and reinforcement fiber, which is coming under predicted fiber class, from the polymer and reinforcement database respectively, based on the design requirements. Fuzzy classification rules to classify fibers into short, medium and long fiber classes are derived based on the fiber length and the computed or derive critical length of fiber. Longitudinal and Transverse module of Polymer Matrix Composite consisting of seven layers with different fiber volume fractions and different fibers' orientations at 0,15,30,45,60,75 and 90 degrees are analyzed through "Rule-of Mixture" material design model. The analysis results are represented in different graphical steps and have been measured with statistical parameters. This data mining application implemented here has focused the mechanical problems of material design and analysis. Therefore, this system is an expert decision support system for optimizing the materials performance for designing light-weight and strong, and cost effective polymer composite materials.

**Keywords:** Knowledgebase, Knowledge Discovery, Fuzzy similarity, Cosine amplitude method, Polymer Matrix Composite.

## 1 Introduction

Polymer matrix composites are lightweight, strong, and energy-efficient materials offer significant advantages to durable-goods manufacturers and to performance-driven markets such as the electronics and communications, computer manufacturing industry and medical fields and are being used in a wide range of applications from aeronautics to sports [3]. A typical Polymer Matrix Composite (PMC) consisting of reinforcing structural constituent and a protective polymer matrix is shown in figure 1.

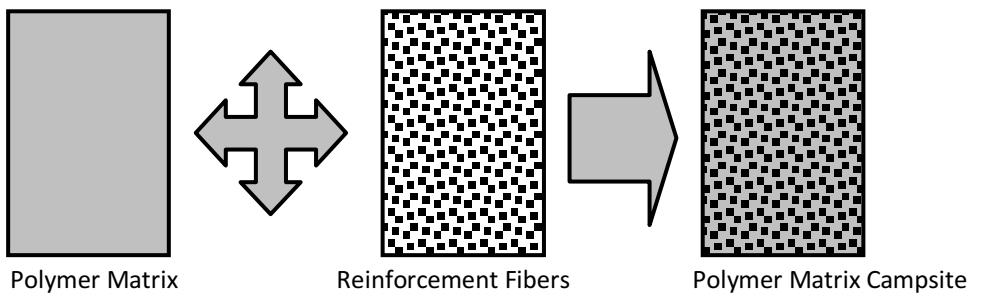

Figure 1: Reinforced Fiber Polymer Matrix Composites Polymer Matrix

The purpose of the reinforcement is to offer specific mechanical solutions in terms of weight, performance and price. The properties of the composite material are significantly better than the sum of the properties of each component, giving materials with high strength-to-weight ratios. As a result, Polymer Matrix Composite parts are generally 20% to 30% lighter than the corresponding metal parts.

PMCs have many benefits to their selection and use. The selection of materials depends on the performance and intended use of the product. The composite designer can tailor the performance of the end product with proper selection of materials based on the end-users requirements.

• *Post-Graduate Department of Studies and Research in Computer Science Mangalore University, Mangalagangotri-574 199, Karnataka, INDIA Ph.No: +91-824-2287670*
*doreswamyh@yahoo.com*

It is important for the end-user to understand the application environment, load performance, durability requirements of the product and convey this information to the composites industry professional. Polymer Matrix Composite design is being played a significant role with the limitations of powerful digital computers in modeling and simulation of Polymer Matrix Composites. Incorporation of knowledge based system and artificial intelligence algorithms in simulation and modeling systems improves the decision making accuracy and design throughput in concurrent manufacturing technology [12].

### 1.1 Back Ground Studies

Over the last two decades, rule based knowledge-based techniques have emerged as powerful decision support tools for simplifying modeling polymer matrix composites. The use of expert systems in the fields of polymer matrix or polymeric-based composites material selection has been reported in the literature [1-15]. Computer packages on the material selection of polymer materials that are currently gaining popularity include Plascams[3] and PERITUS[4]. Plascams[3] is an expert system for plastic material selection, which works using two search routines, which enables the user to search materials qualities from hundreds of materials. PERTUS [4] contains expert systems for selection of polymer as well as for Metal and Ceramic materials, and for design process. It performs manipulations of the data to assist with the preliminary selection of materials.

More emphasis and efforts have been made by the scientists in Intelligent Systems Laboratory (ISL) at Michigan State University, U.S.A., towards developing rule based intelligent decision support systems [5-12] that aid the solution of complex problems through precompiled domain knowledge and specific inferencing techniques. ISL has developed domain rule based intelligent decision support systems such as COMADE [8] for specifying the combinations of polymer matrix materials, chemical agents (curing, reactive diluents), fiber materials, and fiber lengths. The design of polymer composite material systems specifies nothing other than determining valid combinations of material system constituents. Besides the intelligent decision support systems developed at ISL, other researchers have developed generating tools such as Composite Part Designer (CPD) [9], the Composite Designer (COMDES) [10], and Expert Assisted Design of Composite Structures (EADOCS) [11]. These developed tools can share a conceptual design philosophy and have incorporated the capabilities of expert systems of ISL.

Several generic studies on reinforcement fiber performance analysis with polymer matrix have been reported in the literature [14-18]. A Finite element simulation model [14] is developed for the analysis of fiber/matrix interface in unidirectional fiber-reinforced composites and for analyzing stress along longitudinal and transverse direction of fibers. Micromechanical modeling of fiber composites under off-axis loading [15] is implemented in the nonlinear explicit finite element code DYNA3D to demonstrate the effect of the fiber reorientation on the behavior of laminated composite materials in crash simulations. Finite element simulation model [16] is implemented for performance analysis in a material contained 109.8 mm long fiber with uniformly distributed fiber with angle orientations between 11.8 and 56.0.

Data mining application with unsupervised learning SOM algorithm [17] determines the knowledge for clustering reinforcement materials for a candidate polymer based on their effective costs. Knowledge discovery system for cost-effective composite polymer selection implemented with materials design model "Rule-of Mixture"[18] classifies reinforcement fibers into short, medium and long fiber categories and determines the cost effective reinforcement fibers for a given candidate polymer matrix. In this system, critical length of the fiber, $l_c = 0.25mm$, was assumed and the reinforcement fibers under long class were considered for the determination of least cost fibers for the given cylindrical shaft specifications. In the proposed system, the critical length of fibers, which was assumed in knowledge discovery system [18], is computed for the determining the effective performance of polymer matrix composites.

This paper is organized as follows. The second section describes the architecture of the proposed Knowledge Engineering System. The third section describes the polymer matrix selection process and the reinforcement fiber classification on derived fiber critical length. Forth section describes the analysis steps incorporated in Knowledge Engineering Tool (KET) for analyzing the stiffness of the composite when fibers are placed at different orientations. The fifth section describes the experimental results and discussions. Section six gives conclusions and briefs about future work.

## 2. KNOWLEDGE DISCOVERY SYSTEM

"The science of extracting useful information from large data sets or databases" has been rapidly expanding, and attracting many new researchers and users. The underlying reason for such a rapid growth is a great need for systems that can automatically derive useful knowledge from vast volumes of computer data being accumulate and organize. The major trust of research has been to develop a repertoire of tools for discovering both strong and useful patterns in large databases.

Since there is a vast array of different tasks for which knowledge generated from data can be used, a data mining system has to use advanced knowledge representations and be able to generate many different types of knowledge from a given data source. This problem is being partially addressed by the growing inventory of available data mining programs [19-21]. These programs are, however, often arranged into toolboxes, and individuals programs have to be manually invoked. Using such toolboxes can, therefore, be a very laborious and time consuming process, and may require considerable expertise. This problem is being partially addressed by the development of multistrategy data mining systems that integrate different data mining tools [25]. To automate further a data mining process, such tools need to be invokable through a high-level knowledge generation language [24]. Since users want to understand data mining results, an important research direction is also the development knowledge visualization methods [22]. To address the research direction that aims at achieving all the above-mentioned tasks, the term knowledge mining is used here. Knowledge mining can thus be characterized as concerned with developing and integrating a wide range of data analysis methods that are able to derive directly or incrementally new knowledge from large (or small) volumes of data using relevant prior knowledge. The process of deriving new knowledge has to be guided by criteria inputted to the system defining the type of knowledge a particular user is interested in. Algorithms for generating new knowledge must be not only efficient but also oriented toward producing knowledge satisfying the comprehensibility postulate, that is, easy to understand and interpret by the users [23]. Knowledge mining can be simply characterized by the following mapping formula:

**Data + Prior Knowledge + Goal = New Knowledge**

Where Goal is an encoding of the knowledge needs of the user(s), and New Knowledge is knowledge satisfying the Goal. Such knowledge can be in the form of decision rules, association rules, decision trees, conceptual or fuzzy based similarity measure models, mathematical equations, Bayesian nets, statistical summaries, visualizations, natural language summaries, or other knowledge representations.

A typical knowledge engineering system architecture shown in figure 2 is followed the above knowledge mining mapping formula. The proposed system consists of storage repositories: polymer matrix database and reinforcement database, which have organized with data of polymers and reinforcements respectively, knowledgebase of pre designed geometrical attributes of final component design specifications, polymer matrix selection tool for the section of polymer matrix based on the end user requirements, reinforcement fiber classification and selection tool for the classification of reinforcement fibers based on the derived critical length obtained from the attributes of selected polymer matrix and Knowledge Engineering Tool(KET) for engineering knowledge extracted from the tools and knowledge defined in knowledgebase. All the modules are integrated together by user graphical interface that provides designer effective interface environment for polymer matrix composite performance analysis on input and output requirements. The details of procedures implemented in each tools of Knowledge Engineering System are described in the following sub sections.

## 3. POLYMER MATRIX SELECTION

Traditional database techniques have been adequate for many applications involving alphanumeric records, which could be ordered, indexed and searched, for matching patterns in straight forward manner. However, in many scientific database applications, the information is non-numerical by nature. In particular, the large scale materials databases emerge as the most challenging problem in the field of scientific databases. This is due to the large volume of data generated by the powerful computer based modeling and simulation procedures. The database methodologies are concerned with efficient storage and record retrieval. A good database offers fast search coupled with the ability of to handle large verities of queries.

Several computer aided rule based material selection systems [3-4] [8-9] [10-11] have been developed with similarity search techniques to retrieve information from the computerized database on design requirements. Similarity base patterns search involves searching for pattern or objects based on certain characteristics known before hand of the target pattern. Some of the common similarity measures used are Euclidian distance, Linear correlation, Discrete Fourier Transformation etc. The methods described are all mathematical; however in real life most of the properties describing objects or materials are not quantitative, qualitative data including fuzzy data. Fuzzy logic provides the essential tool to utilize qualitative knowledge in the knowledge discovery process [24]. It allows a focused search in the database that can be defined "qualitatively". It also defines the association among materials within the data set which can be expressed in a qualitative format. Fuzzy logic modeling is probability based modeling; it has many advantages [32] over the conventional rule induction algorithms. The first advantage is that it allows processing of very large data sets which require efficient algorithms. Fuzzy logic based rule induction can handle noise and uncertainty in data values well. Fuzzy based procedures [26-28] have been developed for information retrieval from databases.

In the polymer matrix tool proposed in the knowledge engineering system, fuzzy membership functions are implemented for mapping fuzzy data in the input design requirements to crisp values, the fuzzy based Cosine amplitude similarity method [34] is implemented for determining fuzzy similarity strengths between the input requirements and the materials' features stored in the database, and for retrieving the material' features that match the input design requirements.

### 3.1 Fuzzy Membership functions

Fuzzy membership functions are the mapping functions for the fuzzy data containing linguistic terms, fuzzy sets or fuzzy numbers to crisp data. If the fuzzy data are linguistic terms, they are transformed into fuzzy numbers first, and then all the fuzzy numbers are assigned crisp scores. The numerical approximation systems shown in figure 3(a) and 3(b) are proposed for converting linguistic terms in the input design requirements into their corresponding fuzzy numbers.

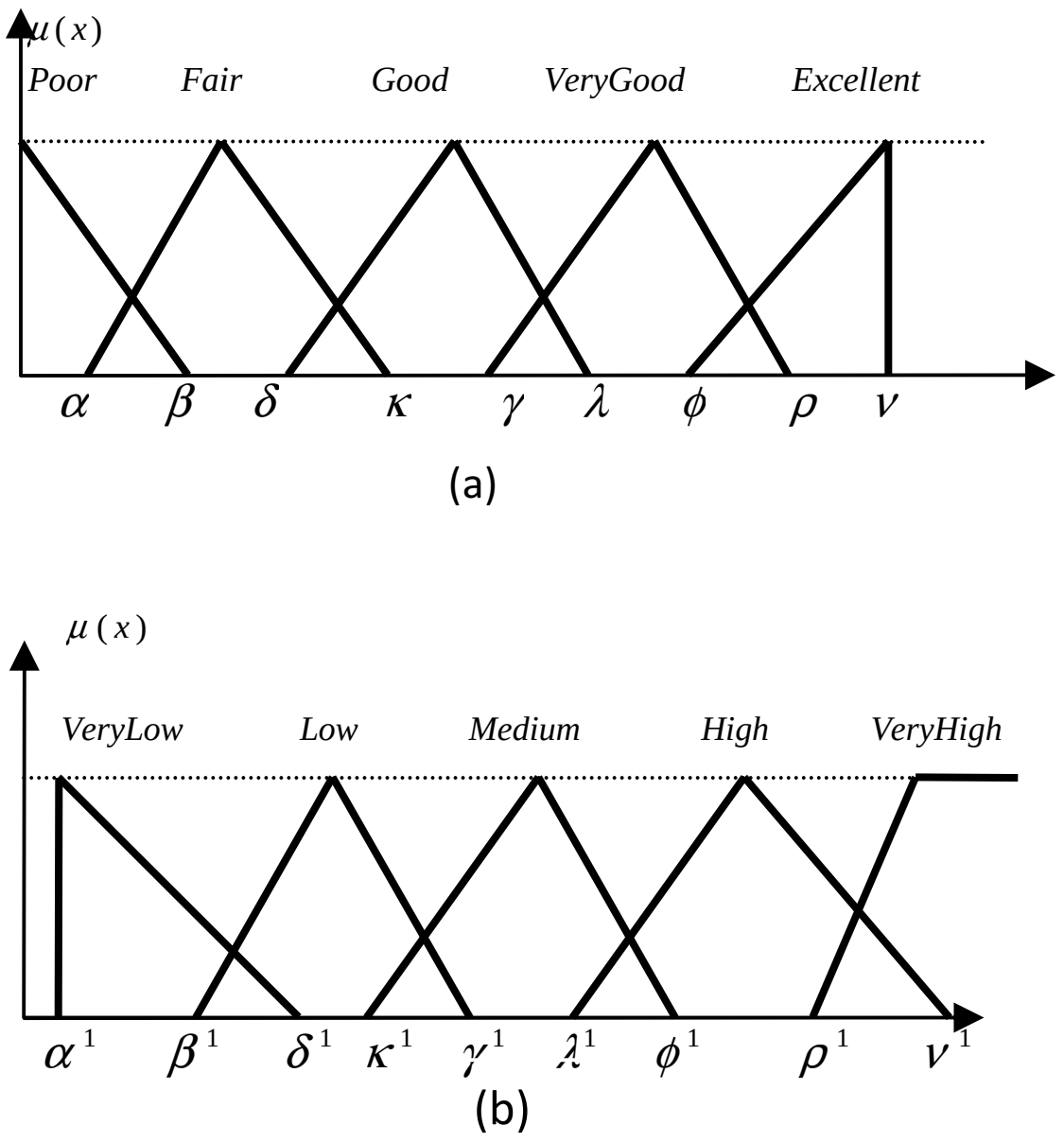

Figure 3: Fuzzy approximation functions for fuzzy input variables.

### 3.2 Fuzzy Similarity Measure

Properties of the materials organized in the database are treaded as N-dimensional feature vectors. The similarity matrix makes use of a collection of N samples and forms a data array X.

$$X = \{X_1, X_2, X_3, X_3, X_5 ........ x_N\} \quad (1)$$

Each of the elements in the data array X is itself a vector of length m that is $X_{i=1..N} = \{x_{i,1}, x_{i,2}, x_{i,3}, x_{i,3}, x_{i,5} ........ x_{i,m}\}$. Hence, each of the data point can be thought of as a point in m-dimensional space, where each sample X needs m coordinates for a complete description. For an input design requirement vector $Y = \{y_1, y_2, y_3, y_4, y_5 ........ y_m\}$, association with data array X is represented by the Cartesian product of two sets Y and Xi is resented by a crisp relation $R_{i=1..N} = \{(Y, X_i) / X_i \subseteq X\}$, where $R_{YXX_{j=1..N}} = \{(y_i, x_{i,j}) / y_i \in Y, x_{i,j} \in X_j\}$. This represents an order pair of $y_i \in Y$ with every $x_{i,j} \in X_j$. The characteristic function

$$\mu_R(y_i, x_{i,j}) = \begin{cases} 1, & (y_{j,}, x_{i,j}) \in R \\ 0, & (y_j, x_{i,j}) \notin R \end{cases}$$

The fuzzy relation, $r_j$, results from pair wise comparisons of two data samples, say $x_{i,j}$ *and* $y_j$, where the strength of the relationship between data sample $x_{i,j}$ and data sample $y_j$ is given by a membership value expressing that strength, that is $r_j = \mu_R(Y_i, X_j)$, The relation matrix will be a vector of size N and as will be the case for all similarity relations. The cosine amplitude method calculates in the following manner and guarantees as do all the similarity methods that $0 \le r_j \le 1$;

$$r_j = \frac{\left|\sum_{k=1}^{m} y_{i,k} x_{j,k}\right|}{\sqrt{\left(\sum_{k=1}^{m} y_{i,k}^2\right)\left(\sum_{k=1}^{m} x_{j,k}^2\right)}} \text{, where } j = 1..N \qquad (2)$$

The close inspection of above equation reveals that this method is related to the dot product for the cosine function. When two vectors are collinear (most similar) their dot product is unity; when two vectors are at right angles to one another (most dissimilar) their dot product is zero. The global similarity measure between the input requirement pattern vector Y and polymer matrix patterns in the data array X is given by

$$S(Y, X_j) = Max\{\mu_{R_1}(Y, X_1) \cup \mu_{R_2}(Y, X_2).....\cup \mu_{R_N}(Y, X_N)\} \qquad (3)$$

### 3.3 Fiber Reinforcement Classification and Selection

Technologically, the most important composites are those in which the dispersed phase is in the form of a fiber. Design goals of fibers-reinforced composites often include high strength and /or stiffness on a weight-basis. These characteristics are expressed in terms of specific strength and specific modulus parameters, which correspond to, respectively, to the ratio of tensile strength to specific gravity and modulus elasticity to specific gravity. Some critical fiber length is necessary for effective strengthening and stiffening of the composite material. The critical length $l_c$ is dependent on the fiber diameter d, fiber tensile strength $\sigma_f$ and on the fiber–matrix bond strength or shear yield strength of the matrix $\tau_c$, accordingly the critical length is defined by

$$l_c = \frac{\sigma_f d}{2\tau_c} \text{ mm} \qquad (4)$$

Reinforcement fibers are classified into short, medium and long fiber classes by fuzzy classification rules derived on fuzzy relationship between the actual length (l) and the computed critical length (lc) of fibers. The derived classification rules defined in figure 4.

$$fiber_length(l) = \begin{cases} short: 0 \quad if \quad fiber_length(l) \le l_c \\ medium\,(fiber_length(l) - l_c)/14l_c \\ \quad if \quad l_c < fiber_length(l) \le 15l_c \\ long: 1 \quad if \quad fiber_length(l) > 15l_c \end{cases}$$

Figure 4: Fuzzy classification rules derived on fiber critical length and fiber length.

Selection of cost effective reinforcement fiber under any fiber length category (Short, Medium and Long) and is associated to a polymer matrix, which has been selected on design requirements in 3.2 is done through the Cosine amplitude similarity method.

## 4. Knowledge Engineering Tool

Designing composite parts offer some challenges and disadvantages. These disadvantages, when known and controlled, can be turned into advantages. When designing composite parts fibers will tend to orient in different directions. This orientation improves mechanical properties in the fiber directions while diminishing in the transverse directions. If the fiber orientation can be predicted and thus controlled, the designer can optimize the geometry and process to produce a lighter weight and low cost product. Current technologies permit fiber orientation to be predicted with molding simulation software [29-31]. Combining results of fibers orientation with structural analysis the molding process can be modified to adapt fiber orientations to strengthen the part in critical structural areas.

The arrangement or orientation of the fibers relative one another, the fiber concentration, and the distribution all have a significant influence on the strength and other properties of fiber- reinforced composites. With respect to the orientation, there are two possible arrangements

1) Parallel alignment of the longitudinal axis of the fibers in a single direction. 2) A totally random alignment. Continuous fibers are normally aligned, where as discontinues fibers may be aligned, randomly oriented or partially oriented. Better over, all composite properties are realized when the fiber distribution is uniform. Material modeling technique -analytical micromechanics model called "Rule of Mixtures" [35] is proposed and implemented in the knowledge engineering tool for the analysis of stiffness of composite part having different planes with uniform distribution of fibers, relative and different fibers' orientations in each plane.

### 4.1 Continuous And Aligned Fibers

The properties of a composite having its fibers aligned are highly anisotropic, that is independent on the directions in which they are measured. The modulus of elasticity of a continuous and aligned fibrous with very good fiber–matrix bond strength or shear yield strength, $\tau_c$, in the direction of alignment or longitudinal direction, is proportional to the sum of products of modulus of elasticity and volume of both polymer matrix and fiber reinforced in the polymer matrix composite, whose volume is equal to Area= Length * Breadth * Height. Polymer Matrix Composite with typical arrangement of continuous and aligned fibers along the direction of the applied load is shown in the figure 5 and its longitudinal modulus of elasticity is

$$E_{CLM} = E_m(1 - V_f) + E_f V_f \text{ GPa.} \qquad (5)$$

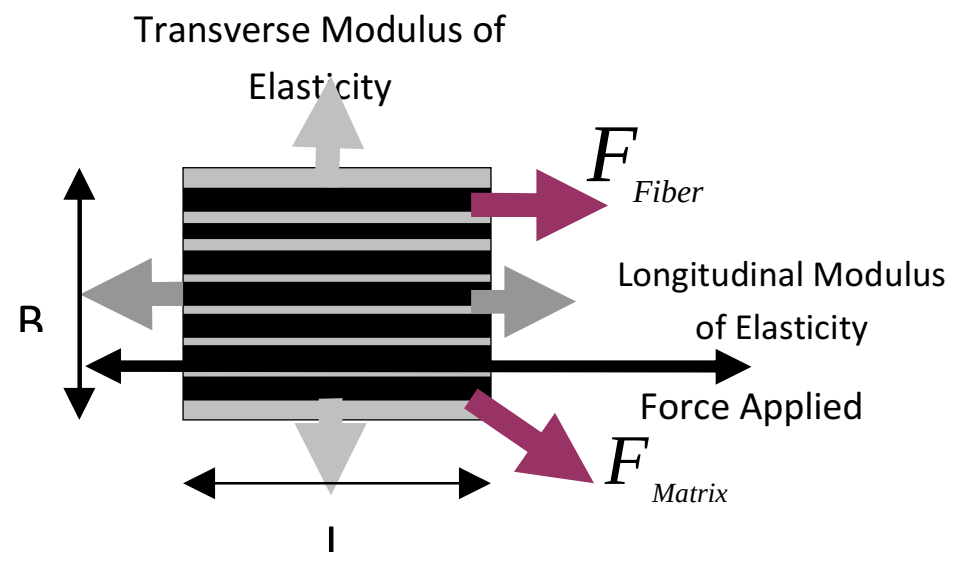

Figure 5: Composite structure having aligned fibers with orientation at $0^0$ along the direction of the force applied.

In a fibrous composite with the applied stress aligned perpendicular to the fibers, the stress is transferred to the fibers through the fiber matrix interface and both the fiber and the matrix experience the same stress. This longitudinal modulus of elasticity of composite sets up a vector relationship between modulus of elasticity and volume of both fiber and matrix phases, whose magnitude satisfies the relational boundary of Cosine amplitude method [35]. A Polymer Matrix Composite having typical fiber arrangements perpendicular to the direction of force applied is shown in figure 6 and its transverse modulus of elasticity is

$$\frac{1}{E_{CTM}} = \frac{v_m}{E_m} + \frac{v_f}{E_f} \text{ GPa} \qquad (6)$$

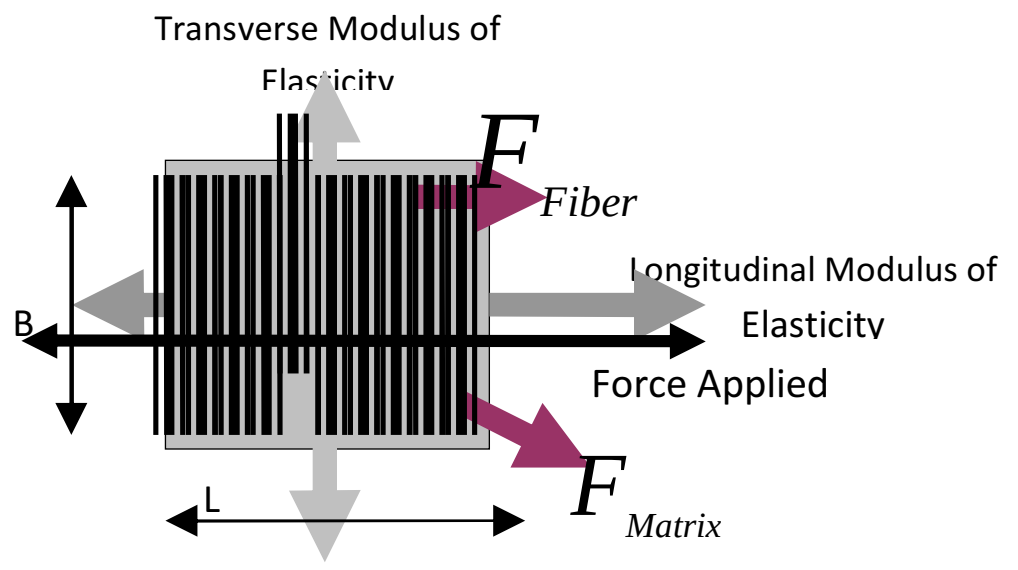

Figure 6: Composite structure having aligned fibers with orientation at 90 along the direction the force applied.

When the fibers are placed at different angles start from 00 -900, the composite stiffness of the continuous and aligned fibers along the direction of the force applied and along the axis of fibers, decreases from the composite stiffness at 00 to the stiffness strength at 900, ie.0 GPa. Mean while the composite stiffness of the continuous and aligned fibers perpendicular to the force applied and along the fibers axis, gradually increases from 0 to 1 GPa. Polymer Matrix Composite in which fibers relatively oriented with $\theta$ = 45 is shown in figure 7 and its composite longitudinal modulus elasticity is

$$E_{CLME} = \cos\theta \ \left(E_m(1-V_f) + E_f V_f\right) \text{ GPa} \qquad (7)$$

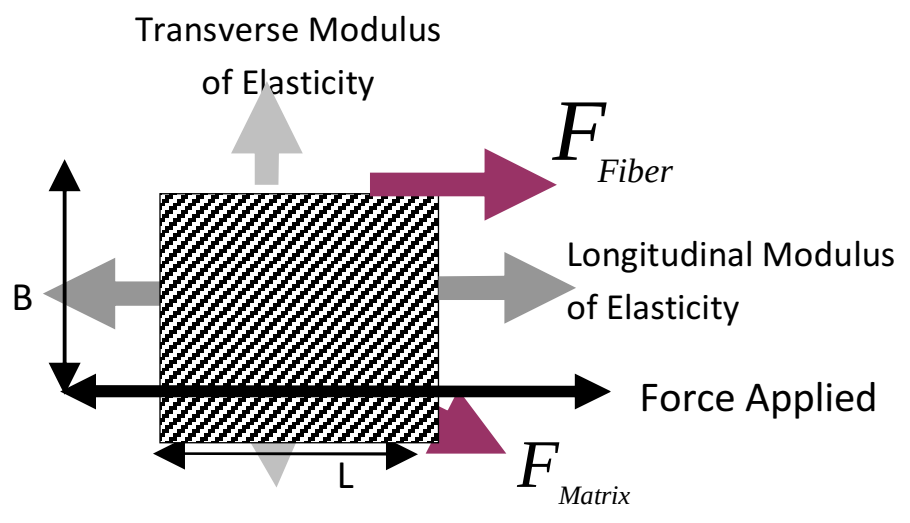

Figure 7: Composite structure having aligned fibers with orientation at $45^0$ along the direction of the force applied.

The transverse modulus of elasticity of composite having continuous fibers oriented at $\theta$ (0-90) with respect to the direction of the force applied is

$$E_{CTME} = (1-Sin\theta)\ \frac{E_f E_m}{(1-V_f)E_f + E_m V_f} \text{ GPa} \qquad (8)$$

### 4.2 Composite Stiffness Analysis

In the composite having volume $V_c = V_f + V_m$, the volumes of fiber $V_f = V_c v_f$ and matrix $V_m = V_c v_m$, are proportional to the volume fractions of fiber, $v_f$ and matrix , $v_m$ respectively such that $v_f + v_m = 1$.

If the fiber phase volume $V_f$ is uniformly distributed into $\{V_f(i), V_f(2), \ldots\ldots V_f(N)\}$ respectively for N planes $\{P_{1,}, P_2, \ldots\ldots P_N\}$, then the fiber volumes $\{V_f(i), V_f(2), \ldots\ldots V_f(N)\}$ corresponding to planes $\{P_{1,}, P_2, \ldots\ldots P_N\}$ are proportional to fiber volume fractions, $\{v(1), v(2), v(3), v(4), \ldots\ldots v(N)\}$.

The volume of a single fiber having fiber length $f_l$ at ith plane is

$$vsf(i) = f_l * v(i) \qquad (9)$$

Number of possible fibers at ith plane is

$$f_n(i) = \frac{V_f(i)}{vsf(i)} \qquad (10)$$

The total volume of fiber phase at ith plane is

$$CV_f(i) = vsf(i) * f_n(i) = V_f(i) \qquad (11)$$

The plane $P_i$ having fiber phase volume, $CV_f(i)$, volume fraction , $v(i)$ ,and if all the fibers , $f_n(i)$ , arranged at $\theta$ degree, the composite longitudinal modulus of elasticity(CLME) composite part is

$$E_{CLME}(P_{i=1..N}, \theta) = \cos\theta \left\lfloor (E_m(V_c - V_f(i)) + E_f V_f(i)) \right\rfloor \text{ GPa} \qquad (12)$$

and composite transverse modulus of elasticity(CTME) of composite part is

$$E_{CTME}(P_{i=1..N},\theta) = (1 - Sin\theta)\left[\frac{E_f E_m}{V_m E_f + E_m V_f(i)}\right] \text{GPa} \qquad (13)$$

A typical composite having N planes and fibers uniformly distributed with relative fiber orientation, $\theta$ in each plane and force acting along the direction of fiber alignment is shown in figure 8.

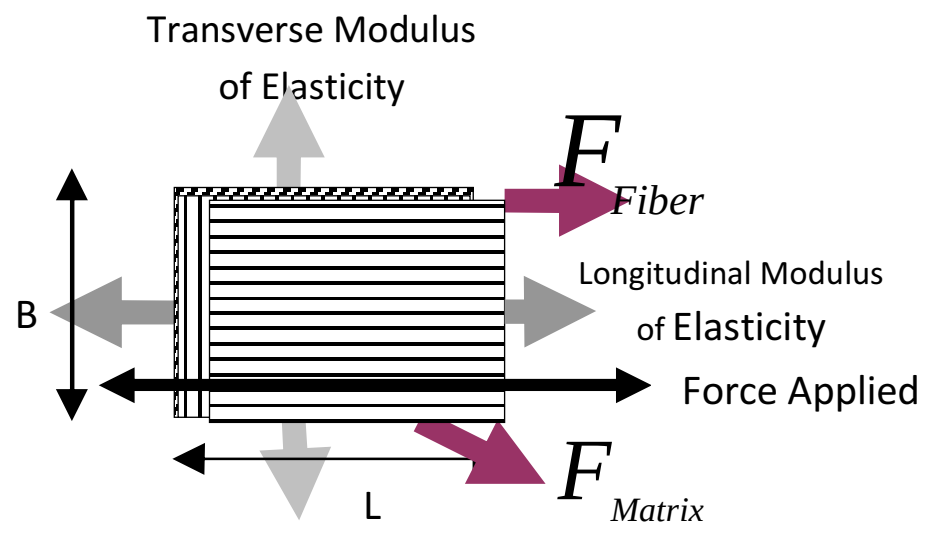

Figure 8: Composite structure having aligned fibers at unique orientation in each plane of the composite.

The mean longitudinal modulus of elasticity (Mean-CLME) the composite having the planes $\{P_{1,}, P_2, ......P_N\}$ in which fibers oriented at $\theta$ degree with respect to the direction of the force applied, is

$$E_{Mean-CLME}(\theta) = \frac{1}{N}\sum_{i=1}^{N}\cos\theta\left[(E_m(V_c - V_f(i)) + E_f V_f(i))\right] \text{GPa} \qquad (14)$$

And the mean transverse modulus of elasticity (Mean-CTME) the composite having planes $\{P_{1,}, P_2, ......P_N\}$ in which fibers oriented at degree $\theta$ with respect to the direction of the fore applied is

$$E_{Mean-CTME}(\theta) = \frac{1}{N}\sum_{i=1}^{N}(1 - Sin\theta)\left[\frac{E_f E_m}{V_m E_f + E_m V_f(i)}\right] \text{GPa} \qquad (15)$$

When the composite consisting of N planes and in each plane, fibers are relatively arranged at $\theta$(i) degree orientation,

Mean-CLME and Mean-CTME are computed by equations (15) and (16) respectively.

$$E_{Mean-CLME} = \frac{1}{N}\sum_{i=1}^{N}\cos\theta(i)\left[(E_m(V_c - V_f(i)) + E_f V_f(i))\right] \qquad (16)$$

$$E_{Mean-CTME} = \frac{1}{N}\sum_{i=1}^{N}(1 - Sin\theta(i))\left[\frac{E_f E_m}{V_m E_f + E_m V_f(i)}\right] \qquad (17)$$

## 5. Experimental Results and Discussion

The knowledge engineering system is implemented for engineering the knowledge/extracting new knowledge from the analysis of the knowledge, which are being extracted by the mining tools on databases and the prior knowledge defined in the knowledgebase, for the desired goals.

The Data mining and knowledge engineering application has projected on mechanical problems related to polymer matrix composites design and their performance analysis in concurrent engineering. The micro structural variables, which will control the properties of the composite, include the mechanical properties of the fiber reinforcement and physical properties such as amount of reinforcement in the matrix, the size and length of the reinforcement and these are depicted in input requirements column in table 2. The properties of the matrix in which the reinforcement is placed include the mechanical properties that are listed in the input requirements columns in the table 1. Properties shown in table 3 are the physical properties of composite that control the mechanical behaviors of both polymer matrix and reinforcement fibers.

Table 1: Input Requirements of Polymer Matrix

| Sl.No | Properties | Numeric and Fuzzy Values |
|---|---|---|
| 1 | Tensile Strength | 57 MPa |
| 2 | Yield Strength | 23 MPa |
| 3 | Elongation | 9.5 |
| 4 | Shear Strength | 43 MPa |
| 5 | Impact Strength | Fair |
| 6 | Modulus of Elasticity | 100 GPa |
| 7 | Creep Strength | Poor |
| 8 | Fatigue Strength | Poor |
| 9 | Density | 0.08 g/cm3 |
| 10 | Melting Point | 750 |
| 11 | Conductivity-Heat | Poor |
| 12 | Conductivity-Electricity | NIL |
| 13 | Thermal Coefficient Expansion(TEC) | Very High |
| 14 | Water Absorption | Poor |
| 15 | Electrical Insulation | Good |
| 16 | Chemical Resistance | Good |
| 17 | Sheet Material | Good |

Table 2: Selected Polymer Matrix : Polyetherimide

| **Sl.No** | **Properties** | **Values** |
|---|---|---|
| 1 | Tensile Strength | 55 MPa |
| 2 | Yield Strength | 23 MPa |
| 3 | Elongation | 10 MPa |
| 4 | Shear Strength | 42 MPa |
| 5 | Impact Strength | Fair |
| 6 | Modulus of Elasticity | 98.99 GPa |
| 7 | Creep Strength | 1.0 |
| 8 | Fatigue Strength | 1.0 |
| 9 | Density | 0.08 g/cm3 |
| 10 | Melting Point | 750 |
| 11 | Conductivity-Heat | 1.0 |
| 12 | Conductivity-Electricity | 0 |
| 13 | Thermal Coefficient | 5.0 |
| 14 | Water Absorption | 1.0 |
| 15 | Electrical Insulation | 3.0 |
| 16 | Chemical Resistance | 3.0 |
| 17 | Sheet Material | 3.0 |

Table 3: Input Design Requirements of Reinforcement Fiber

| | | |
|---|---|---|
| 1 | Diameter | 0.636 mm |
| 2 | Volume fraction | 0.329 % |
| 3 | Length of Fiber | 24.760 mm |
| 4 | Tensile Strength | 3450 MPa |
| 5 | Modulus of Elasticity | 120.0 GPa |

Table 4: Selected Reinforcement Fiber is S-Glass. Length Category : Long type

| | | |
|---|---|---|
| 1 | Diameter | 0.635 mm |
| 2 | Volume fraction | 0.333 % |
| 3 | Length of Fiber | 25.0 mm |
| 4 | Tensile Strength | 3450 MPa |
| 5 | Modulus of Elasticity | 68.69 GPa |

| Table 5: Physical and geometrical properties of composite part | |
|---|---|
| Volume of Composite | ------------ V cm3 |
| Length of Composite | ---------------L cm |
| Breadth of Composite | ---------------B cm |
| Height of Composite | ---------------B cm |
| Number of layers in composite | ---N |
| Size of each layer | - Length* Breadth cm |
| Orientation of fibers | ---------- $\theta_0$ |

### 5.1 Polymer Matrix and Reinforcement Fiber Selection

Selection of polymer matrix from the database containing vast materials' features is an important task. The desired properties of polymer matrix, shown in the table 1, are fed to the polymer matrix selection tool through the interactive interface; firstly it transforms all the fuzzy data into its corresponding crisp data. The fuzzy relationships between input design requirement feature vector and the feature vectors in the polymer database are computed. Finally the feature vector in the polymer database having highest fuzzy similarity strength value is selected and retrieved as the compatible polymer matrix.

The compatible polymer material selected against the input design requirement is shown in the table 2. The fiber class, from which a compatible reinforcement fiber is to be selected, is predicted by the fuzzy classification rules related to fiber length and computed critical length of fiber. Under the predicted fiber class, a compatible reinforcement fiber' futures that match the input design reinforcement requirements in the table 3, are selected and listed in the table 4. The composite properties that control the mechanical behaviors of both polymer matrix and reinforcement fiber are listed in the table 5.

The knowledge extracted from both the polymer matrix and fiber reinforcement tools are tailored for the analysis of the composite fiber performance.

### 5.2 Composite Stiffness Analysis

The mechanical properties of the fiber S-Glass depicted in figure 9 and of the polymer matrix shown in figure 10 are considered the analysis of composite consisting of seven layers and possessing the properties as depicted in table 5. The experimental results were analyzed at different volume fractions of fibers and at different angles of orientations at which fibers placed in composite. The performance analysis results obtained by the knowledge engineering tool for relative fibers' orientations at 0,15,45,60,75 and 90 degree on the planes 1,2,3,4,5,6,and 7 are depicted in the figures 11,12,13,14,15,16,and 17 respectively.

The analysis result obtained for uniform fiber distribution with relative fiber orientation on each plane is depicted in figure 18.

The knowledge extracted on the mechanical behavior of composite stiffness with varying volume fractions are listed as bellow.

Figure 11: Fibers oriented at 0 degree in all the planes. Fibers volume is different at planes. The longitudinal modulus of elasticity increases very fast and the transverse modulus of elasticity also increase to 1very slowly as the volume fraction increases.

Figure 12: Fibers oriented at 15 degree in all the planes. Fibers volume is different at planes. The longitudinal modulus of elasticity relatively lesser at fibers at 0 degree, increases very fast and the transverse modulus of elasticity also increase to 1 very slowly as the volume fraction increases.

Figure 13: Fibers oriented at 30 degree in all the planes. Fibers volume is different at planes. The longitudinal modulus of elasticity relatively lesser at fibers at 15 degree, but increases very fast and the transverse modulus of elasticity also increase to1very slowly as the volume fraction increases.

Figure 14: Fibers oriented at 45 degree in all the planes. Fibers volume is different at planes. The longitudinal modulus of elasticity relatively lesser at fibers at 30 degree, but increases very fast and the transverse modulus of elasticity also increase very slowly as the volume fraction increases.

Figure 15: Fibers oriented at 60 degree in all the planes. Fibers volume is different at planes. The longitudinal modulus of elasticity relatively lesser at fibers at 45 degree, but increases very fast and the transverse modulus of elasticity also increase to 1 very slowly as the volume fraction increases.

Figure 16:

Fibers oriented at 75 degree in all the planes. Fibers volume is different at planes. The longitudinal modulus of elasticity relatively lesser at fibers at 60 degree, but increases very fast and the transverse modulus of elasticity also increase to 1very slowly as the volume fraction increases.

Figure 17: Fibers oriented at 90 degree in all the planes and Fibers' volume is different at planes. The longitudinal modulus of elasticity is approximately equal to zero, while the transverse modulus of elasticity gradually increases from 0 to 1

Figure 18: Fibers are university distributed with relative orientation in each plane, but fiber volume is different and orientation fiber volume is different in different plane. The longitudinal modulus of elasticity decreases to zero as the volume fraction of fiber increases with different angle of orientations. Mean while the transverse modulus of elasticity increases from 0 to 1 gradually.

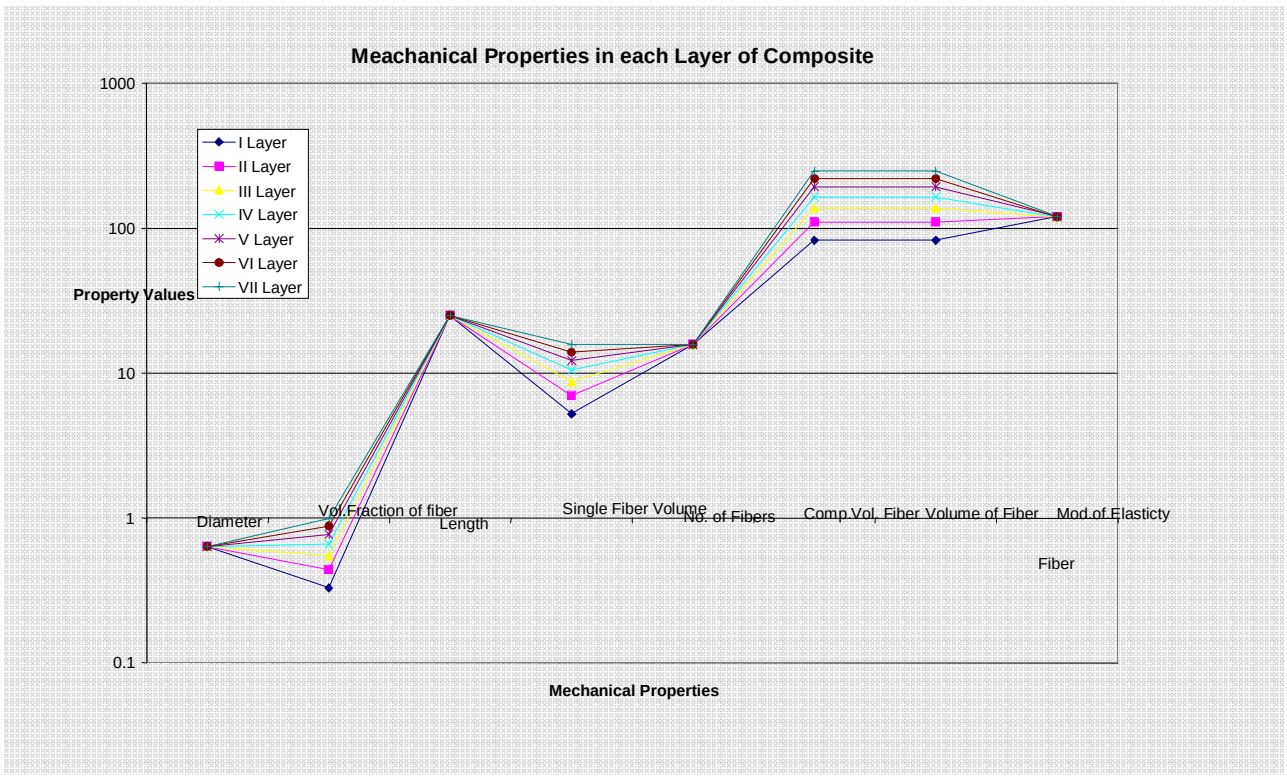

Figure 9: Mechanical Properties of fiber S-glass having diameter d = 0.635, volume varying volume fraction from 0.33-0.99, length = 25mm, Modulus of Elasticity = 120 GPa

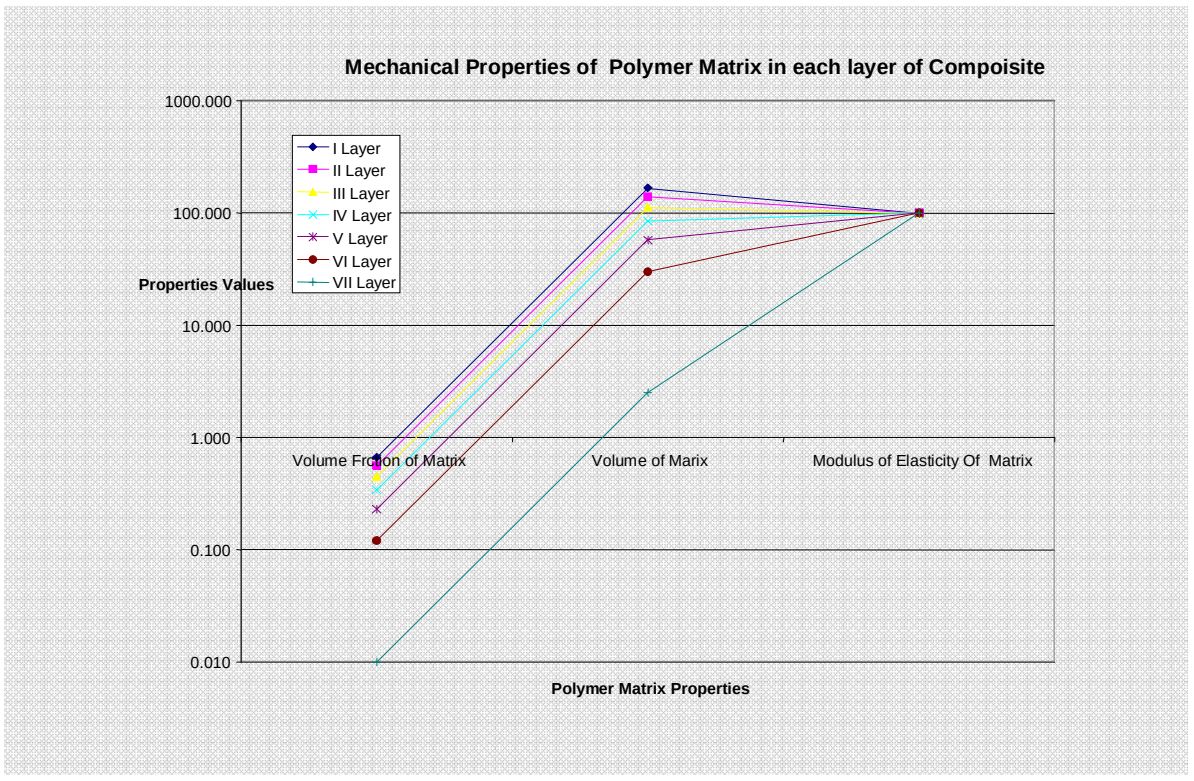

Figure 10: Mechanical properties of Polymer Matrix Polyetherimide having volume fraction varying from 0.67-0.01, Volume of Matrix varying from 167.5 to 2.50 and Modulus of Elasticity = 100

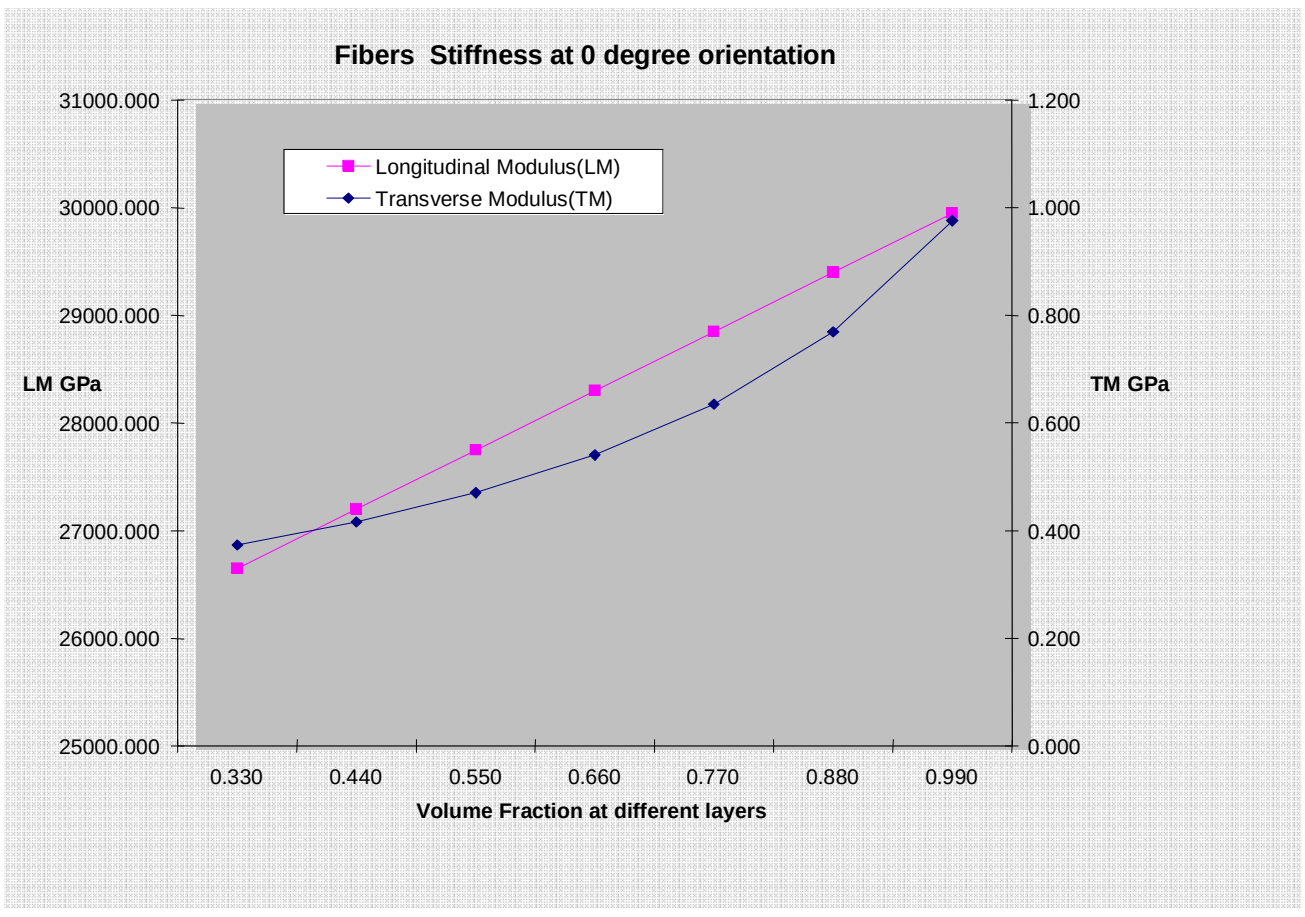

Figure 11: Longitudinal modulus of elasticity and Transverse modulus of elasticity of composite with fibers orientation at 0 degree.

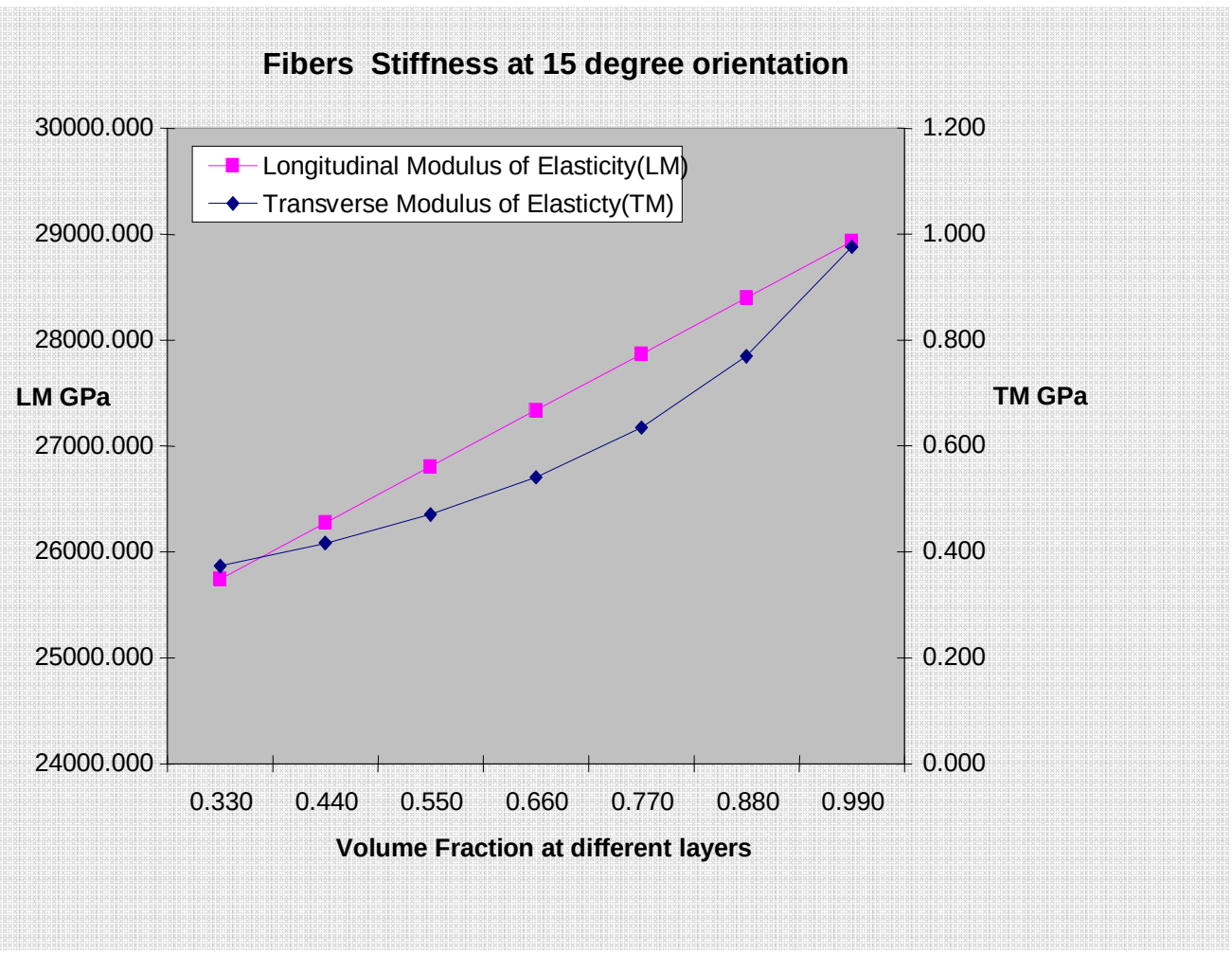

Figure 12: Longitudinal modulus of elasticity and Transverse modulus of elasticity of composite with fibers orientation at 15 degree.

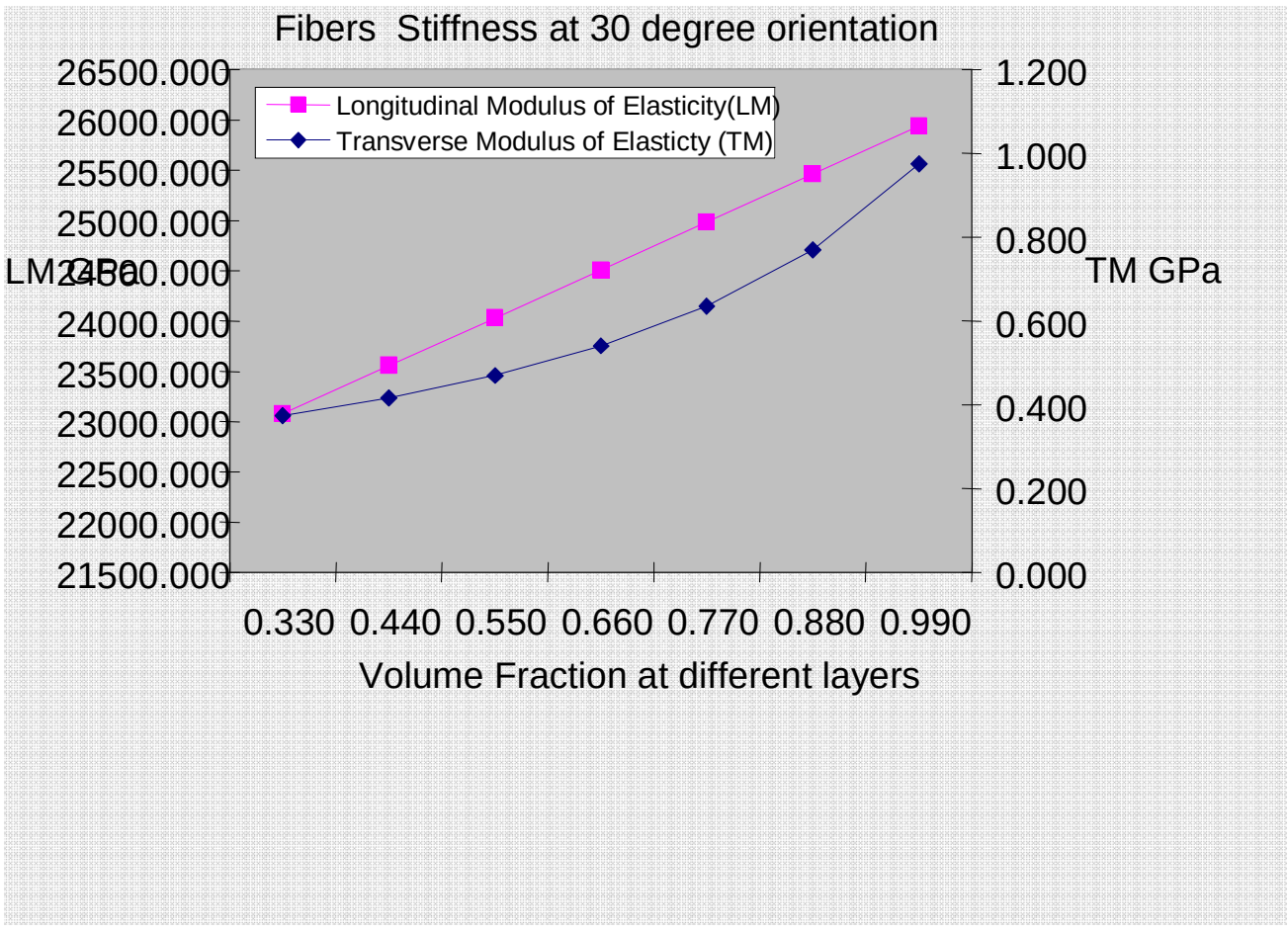

Figure 13: Longitudinal modulus of elasticity and Transverse modulus of elasticity of composite with fibers orientation at 30 degree.

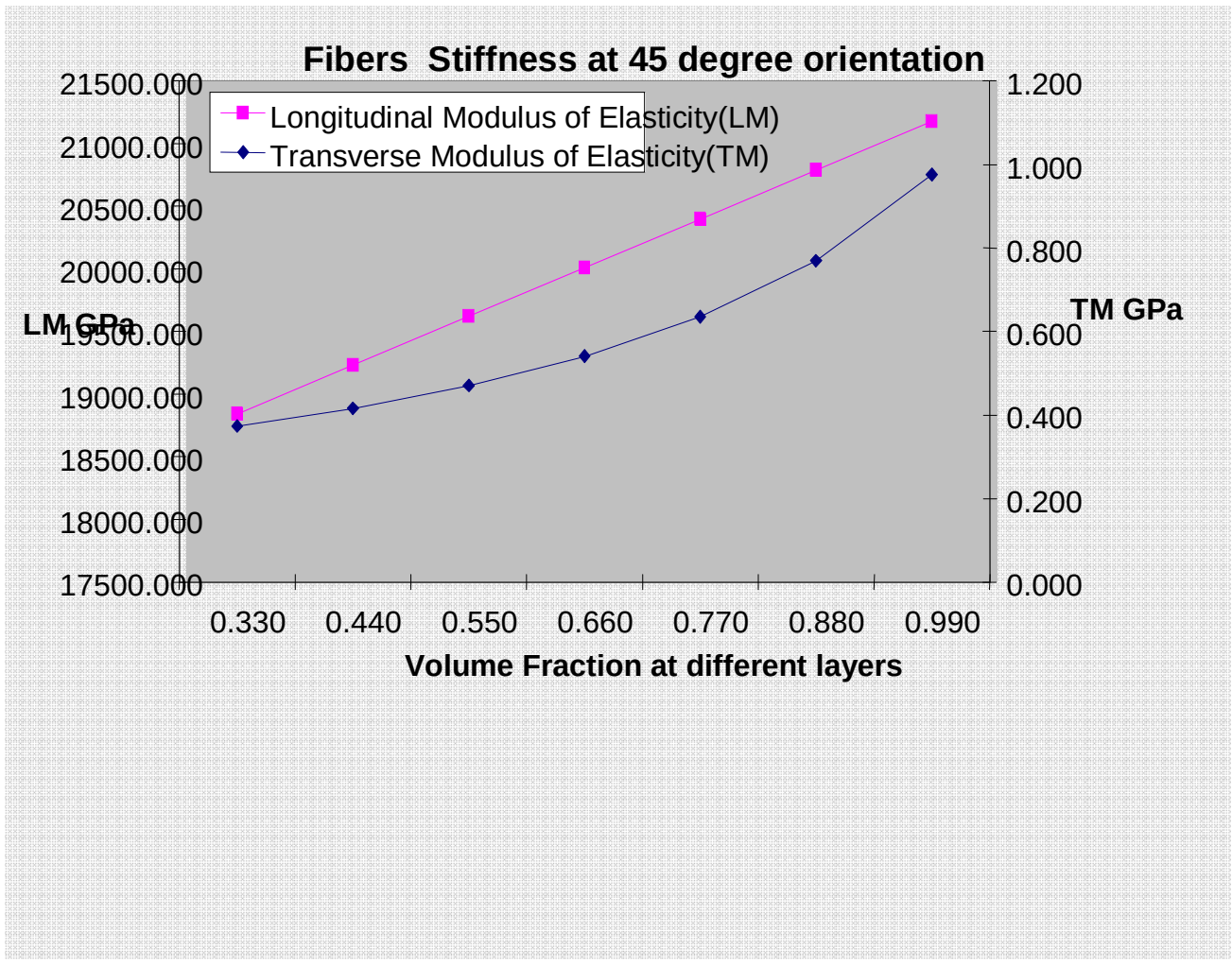

Figure 14: Longitudinal modulus of elasticity and Transverse modulus of elasticity of composite with fibers orientation at 45 degree.

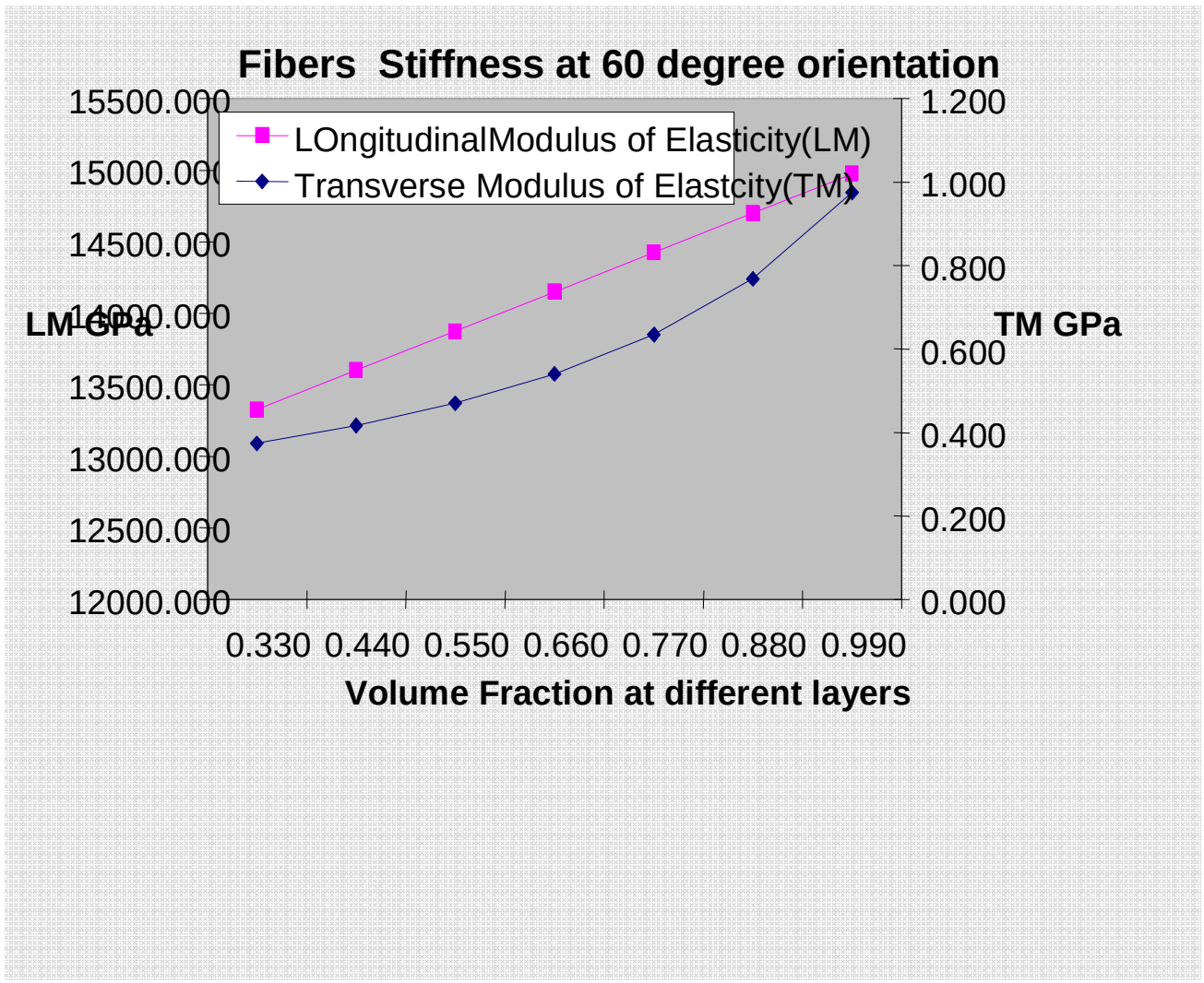

Figure 15: Volume fractions different at different layers with unique fiber orientation at 60 degree increase longitudinal modulus and decrease transverse modulus.

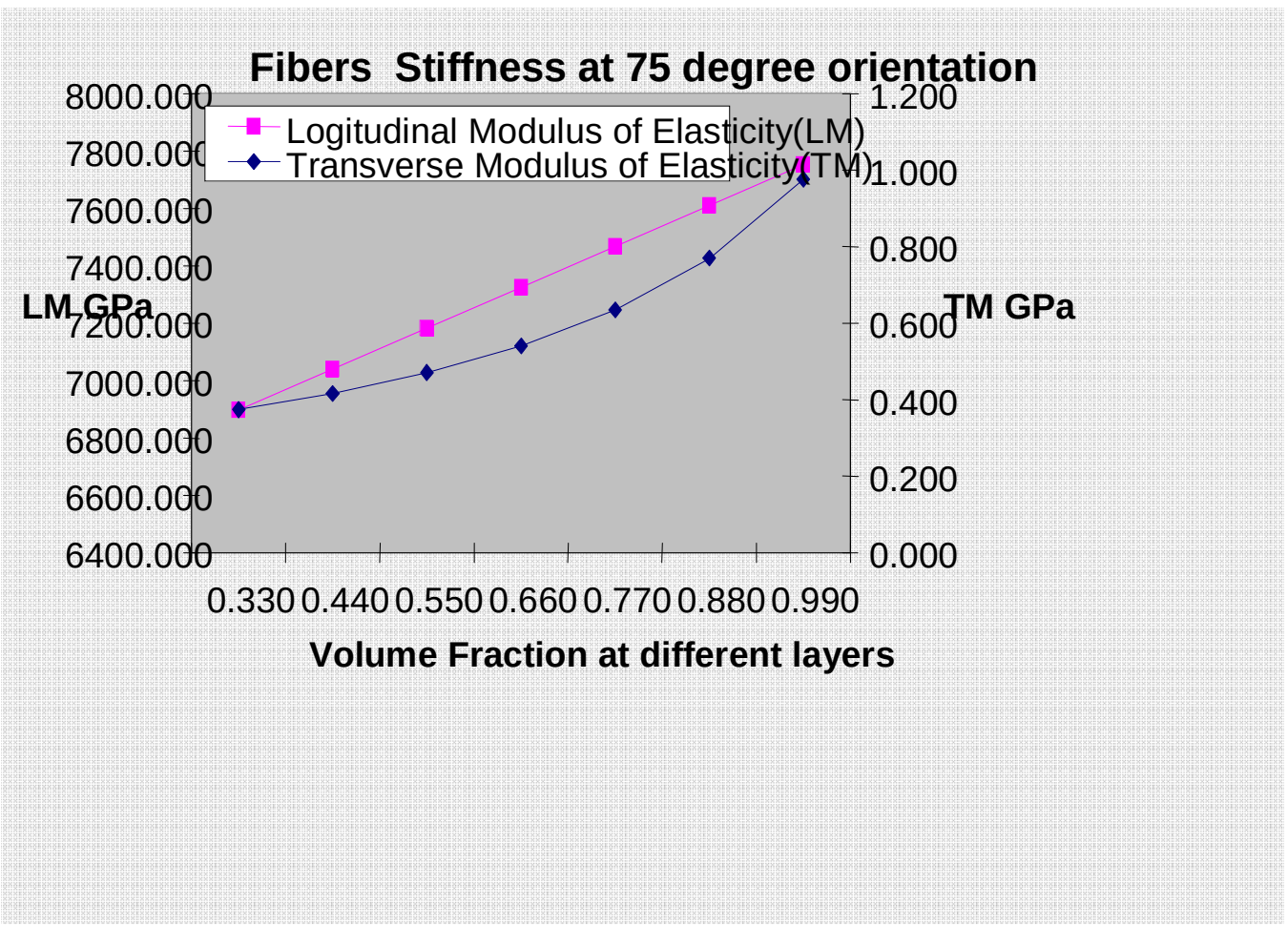

Figure 16: Volume fractions different at different layers with unique fiber orientation at 75 degree increase longitudinal modulus and decrease transverse modulus.

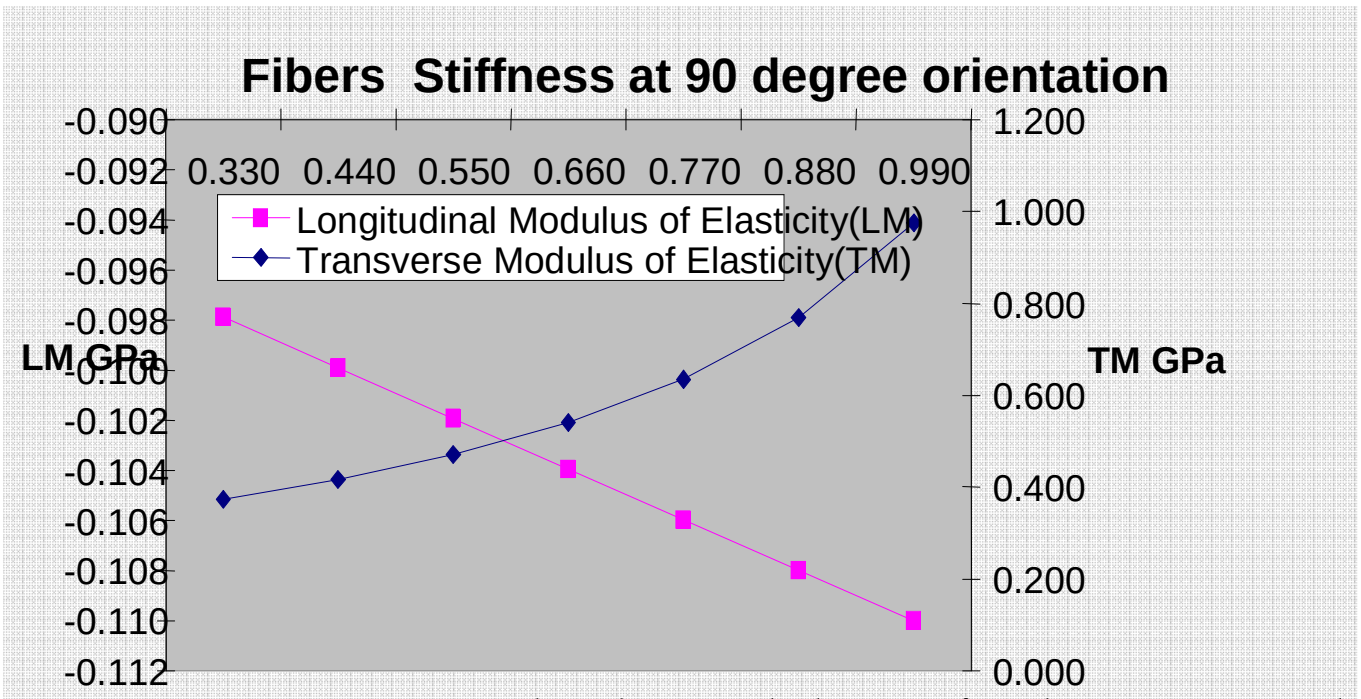

Figure 17: Longitudinal modulus of elasticity and Transverse modulus of elasticity of composite with fibers orientation at 90 degree.

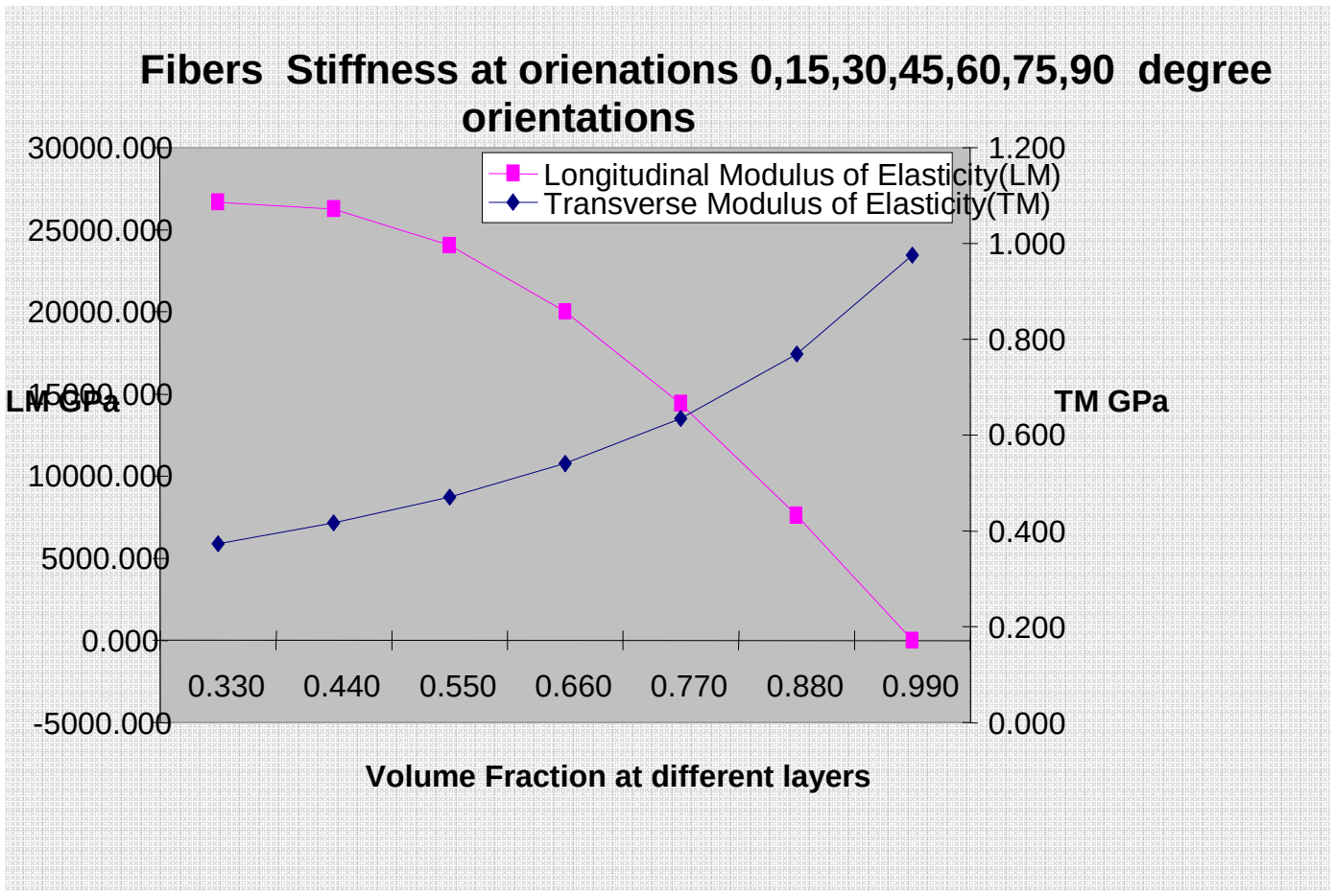

Figure 18: Longitudinal modulus of elasticity and Transverse modulus of elasticity of composite with fibers orientations at 0, 15, 30, 45, 60, 75, and 90 degrees

### 5.3 Mean stiffness Analysis

The mean longitudinal stiffness of the fiber reinforcement composite decreases from to zero as the fiber volume increases with changing relative fiber orientations each plane of the campsite. Table 6 describes the statistical measures of the mean composite stiffness.

Table 6: Sowing the statistical measures of composite stiffness when all the fibers uniformly distributed with relative orientations and with mean volume fraction of fiber = 0.66, mean volume of single fiber = 10.477577, mean volume fiber phase = 165.0 ,mean of volume fraction of matrix = 0.36, VC = 7 * 250

| Composite No. | Orientations | Mean -CLME | Mean -CTME |
|---|---|---|---|
| 1 | 0 | 28300 | 0.59757143 |
| 2 | 15 | 27335.6964 | 0.59757143 |
| 3 | 30 | 24508.5017 | 0.59714286 |
| 4 | 45 | 20011.085 | 0.59757143 |
| 5 | 60 | 14149.9399 | 0.597 |
| 6 | 75 | 7324.49529 | 0.59771429 |
| 7 | 90 | -0.104 | 0.597 |

Table 7 describes the statistical measures of the behavior of the composite longitudinal modulus and composite transverse modulus of elasticity, when fibers are uniformly distributed and different fiber orientations in different planes.

Table 7: Showing the mean of both longitudinal and transverse modulus composite containing seven layers corresponding to fibers oriented at 0, 15, 30,45,60,75 and 90 respectively.

Diameter of Fiber d = 0.635mm, Length = 25.00 cm, Total Fiber Volume = 15.748, Modulus of Elasticity = 120GPa, Polymer Matrix : Volume of Matrix = 100Ccm3, Modulus of Elasticity = 100 GPa, Fiber orientation degree = 0: Composite Volume(VC) = 7 * 250cm3

| Fiber | | | | Matrix | | Composite Modulus | |
|---|---|---|---|---|---|---|---|
| Fiber Orientations | Vol. fraction of Fiber | Vol. of Single Fiber | Vol. Fiber Phase | Vol. fraction of Matrix | Matrix Phase Vol. | Mean -CLME | Mean -CTME |
| 0 | 0.33 | 5.239 | 82.5 | 0.67 | 167.5 | 26650 | 0.374 |
| 15 | 0.44 | 6.985 | 110 | 0.56 | 140 | 26273.178 | 0.417 |
| 30 | 0.55 | 8.731 | 137.5 | 0.45 | 112.5 | 24032.188 | 0.47 |
| 45 | 0.66 | 10.478 | 165 | 0.34 | 85 | 20011.085 | 0.541 |
| 60 | 0.77 | 12.224 | 192.5 | 0.23 | 57.5 | 14424.939 | 0.635 |
| 75 | 0.88 | 13.97 | 220 | 0.12 | 30 | 7609.193 | 0.77 |
| 90 | 0.99 | 15.716 | 247.5 | 0.01 | 2.5 | -0.11 | 0.975 |
| Sum | 4.62 | 73.343 | 1155 | 2.38 | 595 | 119000.473 | 4.182 |
| Mean | 0.66 | 10.47757 | 165 | 0.34 | 85 | **17000.0676** | **0.59742857** |

## 6. Conclusions and Future Work

### 6.1 Conclusions

In this paper, Knowledge Mining and Knowledge Engineering tools are proposed and implemented for Engineering knowledge extracted from the mining tools. The knowledge engineering models proposed and implemented in the system are tested with the polymer matrix, reinforcement fiber and composite properties as depicted in table 1, 3 and 5 respectively. The stiffness of the composite is analyzed when the fibers having different volume fractions and placed at different orientations in the composite material. Knowledge extracted from the Knowledge Engineering System are listed as below

1. Compatible polymer matrix selected from the polymer database shown in the table 2.
2. Compatible reinforcement fiber selected from the fiber database.
3. Knowledge about the fiber class- whether short, medium or long fibers. Fiber.
4. From the figures 12-16, it is found that longitudinal modules of elasticity increases from the first plane to last plane as the volume fraction increases.
5. Figure 17 depicts that as the fiber volume faction increases and fiber orientation changes, the stiffness of the composite decreases.
6. The longitudinal stiffness of the composite increases as the volume fiber phase increase. The he means longitudinal stiffness of all the composite planes decreases as the orientations of the fibers placed in the matrix phase changes from 0-90 degree.
7. The mean transverse stiffness of the composite is constant for the mean volume fraction, and mean fiber volume and is independent of the fibers orientations in each plane.
8. The transverse modulus of elasticity of composite increases in all the cases within the interval (0, 1) and is negligible compared to longitudinal modules in all the cases except fibers' orientation at 90 degree.
9. From the figure 18 it is found that in a composite, if each layer contains fiber at specific orientation $\theta(i)$ and orientations of fibers at different plane are different, the equilibrium longitudinal of modulus and transverse modulus are approximately equal to $E_{Mean-CLME}$ = 17000.0676 and $E_{Mean-CTME}$ 0.59742857 respectively.
10. This system can be generalized for analyzing the stiffness of various composite materials by defining their different geometries in the knowledgebase of this system.
11. This system can also be generalized for analyzing the stiffness of other composite materials like fiber reinforcement metal matrix, and fiber reinforcement ceramic composite materials.

### 6.2 Future works

i. The future work is focused on developing generalized software tools for the proposed knowledge engineering system.
ii. Designing and implementation of a mathematical model for the determination the equilibrium state of the composite stiffness where both longitudinal and transverse moduli are balanced.
iii. Designing data mining algorithms for the prediction of properties of both polymer matrix and reinforcement fiber, when the optimal performance parameters of a desired composite are given – using neural network approach.

### References

[1] S. M. Sapuan, "A computer-aided material selection for design of automotive safety critical components with novel materials", Malaysian Journal of Computer Science, Vol.12. no.2, December 1999, pp.37-45.

[2] A.M. Yaacob, M. Ahmad, K.Z.M. Dahlan and S.M. Sapuan, "Experimental studies on fiber orientation of short glass fiber reinforced injection molded thermoplastic composites," Proceedings of Advanced Materials Conference, Advanced Technology Congress, Putra Jaya, 20-21 May 2003 (CD-Rom Proceedings).

[3] C. Maier, Materials Witness, "British Plastic and Rubber," April 1993.

[4] G S Dodd and A G Fairfull,"Knowledge-Base systems in Materials Selection, In B.F Dyson and D R Hayhurst (Eds), Material and Engineering Designs; in the next decade, 1999. pp 129-131.

[5] Timothy Lenz ,James K. McDowell, Ahmed Kamel,Jon Sticklen and Martin C. Hawley ,"A Decision Support Architecture for Polymer Composites Design: Implementations and Evolution," Proceedings of the Eleventh Conference on Artificial Intelligence

for Applications, pp. 102-108, IEEE Computer Society Press, February 20-22 1995.

[6] Timothy J. Lenz, James K. McDowell, Martin C. Hawley, Ahmed Kamel, and Jon Sticklen, "The Evolution of a Decision Support Architecture for Polymer Composites Design," IEEE Expert: Intelligent Systems and Their Applications, vol. 11, no. 5, pp. 77-83, Oct., 1996.

[7] James K. Mcdowell, Ahmed M. Kamel, Jon Sticklen, and Martin C. Hawley, "Integrating Material/Part/Process Design for Polymer Composites: A Knowledge-Based Problem-Solving Approach," Journal of Thermoplastic Composite Materials, Vol. 9, No. 3, 218-238 ,1996.

[8] Lenz, T., McDowell, J. K., Moy, B., Sticklen, J., and Hawley, M. C. "Intelligent Decision Support for Polymer Composite Material Design in an Integrated Design Environment.", In American Society of Composites 9th Technical Conference on Composite Materials, pp. 685-691, 1994. Newark, Delaware.

[9] Michaeli, W., and A. Biswas., "Knowledge- and Case-based Computer Support for the Design of Composite Parts", Proceedings of the ANTEC '94, San Francisco. 2347-2350,1994.

[10] Edwards, K. L., P. M. Sargent, and D. O. Ige. "Designing with Composite Materials - An Expert System Approach", in the proceedings the American Society of Mechanical Engineers 2nd Biennial European Conference on Engineering Systems Design & Analysis, London, 1994.

[11] Netten, B. D., R. A. Vingerhoeds, and H. Kippelaar. "Expert Assisted Conceptual Design: An Application to Fiber Reinforced Composite Panels", In R. M.Oxman, M. F. T. Bax, & H. H. Achten (Eds.), Design Research in the Netherlands, pp. 125-139, Eindhoven: Faculty of Architecture, Planning, and Building Science. Eindhoven University of Technology, 1995.

[12] Moy, B., McDowell, J. K., Lenz, T. J., Sticklen, J., & Hawley, and M. C. "Integrated Design and Agile Manufacturing in Polymer Composites. The Role of Intelligent Decision Support Systems.", In Intelligent Manufacturing Systems Workshop, 14th International Joint Conference on Artificial Intelligence, IJCAI 95, .Montreal, CA:1995.

[13] T. J. Lenz, M. C. Hawley, and Jon. Sticklen, "Virtual Prototyping in Polymer Composites", Journal of Thermoplastic Composite Materials, 11, 1998.

[14] Wen S. Chan and Lee Ann Johnson, "Analysis of Fiber/Matrix Interface in Unidirectional Fiber-Reinforced Composites, Journal of Thermoplastic Composite Materials, Vol. 15, No. 5, 389-402,2002.

[15] Ala Tabiei, and Ivelin Ivanov, "Fiber Reorientation in Laminated and Woven Composites for Finite Element Simulations", Journal of Thermoplastic Composite Materials, Vol. 16, No. 5, 457-474, 2003.

[16] T. S. Creasy and Y. S. Kang,"Fiber Orientation during Equal Channel Angular Extrusion of Short Fiber Reinforced Thermoplastics", Journal of Thermoplastic Composite Materials, Vol. 17, No. 3, 205-227, 2004.

[17] Doreswamy, S. C. Sharma, M Krishna, and H N Murthy` "Data Mining Application In Knowledge Extraction Of Polymer And Reinforcement Clustering", International Conference on Systemics, Cybernetics And Informatics, ICSCI-2006, January 4th to 8th, 2006, Pentagram Research Center, Hyderabad, INDIA. Pp.no.562-566.

[18] Doreswamy, S. C. Sharma and M Krishna, "Knowledge Discovery System for Cost-Effective Composite Polymer Selection-Data Mining Approach",12th International Conference on Management of Data COMAD 2005b, December 20-22, 2005 - Hyderabad, INDIA, pp.no. 185-190.

[19] Agrawal, R., Mannila, H., Srikan R., Toivonen, H., and Verkamo, A. I., "Fast Discovery of Association Rules, in Advances in Knowledge Discovery and Data Mining," U.M. Fayyad, G. Piatetsky-Shapiro, P., Smyth, and Uthurasamy, R. (Eds.), AAAI Press, Menlo Park, CA. 1996.

[20] Breiman, L., Friedman, J.H., Olshen, R.S., and Stone, C.J., "Classifications and Regression Trees," Wadsworth Statistical Press, Belmont, CA, 1984;

[21] Witten, I. H., Moffat, A., and Bell, T.C., "Managing Gigabytes: Compressing and Indexing Documents and Images", 2nd ed., Morgan Kaufman, San Francisco, CA, 1999.

[22] Kevin M. Dillon and Patrick J. Talbot, W. Daniel Hillis, "Knowledge Visualization: Redesigning the Human-Computer Interface", Technology Review Journal Spring/Summer 2005. Available at the internet address: http://www.idemployee.id.tue.nl/g.w.m.rauterberg/amme/dillon-et-al-2005.pdf.

[23] Michalski, R. S., "A Theory and Methodology of Inductive Learning, in Machine Learning: An Artificial Intelligence Approach," R. S. Michalski, J. Carbonell and T. Mitchell (Eds.), pp. 83-134, Morgan Kaufman Publishing Co., Palo Alto, 1993.

[24] Michalski, R.S. and Kaufman, K.A., "Data Mining and Knowledge Discovery: A Review of Issues and a Multistrategy Approach," In Machine Learning and Data Mining: Methods and Applications, Michalski, R.S., Bratko, I. and Kubat, M. (eds.), London, John Wiley & Sons, pp. 71-112, 1998.

[25] Morik, K. and Brockhausen, P., "A Multistrategy Approach to Relational Knowledge Discovery in Databases," Proceedings of the third International Workshop on Multistrategy Learning (MSL-96), pp. 17-27, 1996.

[26] M.J. Swain and D.H Ballard, "Color Indexing International," Journal of Computer Vision, Vol.7.No.1. Pp.11-32, 1991.

[27] Rami Zwick, "Measure of Similarity among fuzzy concepts: a comparative analysis," International Journal of approximate reasoning, pp.221-242,1987.

[28] R. Jain, S.N.J. Murthy, and L. Tran," Similarity measures for Image databases", IEEE Transaction on PAMI.Vol.17, No.7, pp.1247-1254, 1995.

[29] Osswald, T.A. E.M. Sun and S.C Tseng, "Orientations and Warpage Predictions in Polymer Processing," a chapter innovation in polymer processing: Molding, edited by J.F Stevenson Hanser, 1996.

[30] Davis.B.A., R.P.Theriaull, T.A,Osswald, "Optimization of the Compression/ Injection Molding process using Numerical Simulations.", ASME Conference,1997.

[31] Rios A., B.Davis, P.Gramann, "Computer Aided Engineering in Compression Molding", Composite Fabricators Association annual conference, 2001.

[32] Maeda, H. Ashida, Y. Taniguchi, and Y. Takahashi, "Data mining system using fuzzy rule induction," Proc. IEEE Int. Conf. Fuzzy Syst. FUZZ IEEE 95, pp. 45--46, Mar. 1995.

[33] W. Pedrycz. "Data mining and fuzzy modeling," In Proc. of the Biennial Conference of the NAFIPS, pages 263--267, Berkeley, CA, 1996.

[34] Timothy J. Ross, "Fuzzy Logic with Engineering Applications," MCGraw-Hill Inc., 1997.

[35] P.K.Valavala and G.M.Odegard, "Modeling Technique for Determination of Mechanical properties of Polymer Nanocomposites", Advance Material Science.Vol.9.34-35, 2005.